\documentclass[11pt,a4paper]{article}
\usepackage[hyperref]{eacl2021}
\usepackage{times}
\usepackage{latexsym}
\usepackage{tabularx}

\usepackage{microtype}
\usepackage{graphicx}

\usepackage{amsmath}
\usepackage{float}
\newcommand{\minus}{\scalebox{0.75}[1.0]{$-$}}

\aclfinalcopy % Uncomment this line for the final submission
\title{Injecting Hierarchy with U-Net Transformers}
\author{David Donahue, Vladislav Lialin, Anna Rumshisky\\
Text Machine Lab\\
Dept. of Computer Science\\
UMass Lowell\\
\{david\_donahue, vladislav\_lialin\}@student.uml.edu, arum@cs.uml.edu \\
}

\begin{document}

\maketitle
\begin{abstract}
The Transformer architecture has become increasingly popular 
over the past two years,
%over the past couple of years, 
owing to its impressive performance on a number of natural language processing (NLP) tasks.
However, all Transformer computations occur at the level of word representations and therefore, it may be argued that Transformer models do not explicitly attempt to learn hierarchical structure which is widely assumed to be integral to language.
In the present work, we introduce hierarchical processing into the Transformer model, taking inspiration from the U-Net architecture, popular in computer vision for its hierarchical view of natural images.
We empirically demonstrate that the proposed architecture outperforms both the vanilla Transformer and some strong baselines in the domain of chit-chat dialogue.

\end{abstract}

\section{Introduction}

The recently introduced Transformer architecture \cite{vaswani2017attention} and its variants have achieved state-of-the-art performance in a number of tasks including machine translation, language modeling \cite{dai2019transformer}, question answering \cite{lan2019albert}, and others, with Transformer-based models dominating leaderboards for many common benchmarks \cite{wang2018glue,wang2019superglue,lai2017race}. 

The Transformer removes the need for sequential computation of the input sequence and instead employs a global self-attention mechanism to allow distant token positions to exchange information in a constant number of steps. In short, it 
%the Transformer
operates exclusively through token-level interactions to produce contextualized embedding representations at the encoder and decoder output layers. This attention mechanism is credited with the Transformer's continued success in %a variety of 
many NLP tasks.

While the Transformer has been applied most extensively in machine translation, it has also recently been applied successfully in  dialogue generation~\citep{Wolf2019TransferTransfoAT,Budzianowski2019HelloIG}. Conversational dialogue can be thought of as a conditional language modelling task, where an output response needs to be produced in the presence of an input conversation history. In real-world dialogue, the conversation history can be extensive, and often an effective conversationalist must interpret both the low-level details and the high-level trajectory of the conversation, including topic structure and sentiment.
%(e.g. conversation sentiment and topic). 
%
While the Transformer explicitly targets capturing word-level interactions through the attention mechanism, 
%in the context of dialogue, 
learning a proper representation at different levels of abstraction may pose a challenge for such models, 
since they lack an explicit mechanism for learning hierarchical representations.

In this paper, we propose a novel U-Net Transformer architecture to address this issue.
%\footnote{Codebase will be released upon publication.}.
%the issues outlined above\footnote{Codebase will be released upon publication.}.  
%In the present work, 
We introduce hierarchical processing into the Transformer model by taking inspiration from the U-Net architecture, popular in computer vision for its hierarchical view of natural images.  The U-Net architecture has produced recent state-of-the-art performance in semantic segmentation \cite{chen2018encoder}, image generation \cite{song2019generative}, and unsupervised alignment \cite{zhu2017unpaired}.

We empirically demonstrate that the proposed architecture outperforms several strong baselines, including a vanilla Transformer, on conversational dialogue generation.
% and machine translation benchmarks.
%
Using the Cornell Movie Dialogue and PersonaChat datasets
% , as well as the WMT2014 English-German dataset,
our model achieves a consistent drop in perplexity
% and increase in the BLEU score
over the best baseline performance.
%Our model demonstrates a consistent improvement over the best baseline performance for the Cornell Movie Dialogue and PersonaChat datasets, as well as the WMT2014 English-German dataset, 

% In the following sections, we motivate the proposed architecture and describe its construction and operation in a step-wise fashion. We then describe conceptually how the proposed U-Net Transformer addresses the pitfalls of the Transformer models described above and report on empirical evaluation of the proposed architecture against multiple baselines. 

%In the next section, we will discuss the Transformer and U-Net architectures, along with relevant works motivating our proposed architecture. In the following section (Model Description), we describe the construction and operation of the U-Net architecture in a step-wise fashion. Throughout the remainder of the paper, we will describe the evaluation of our approach to combining the Transformer and U-Net architecture into a single model. We will describe conceptually how the novel U-Net Transformer addresses the pitfalls in the Transformer above and perform multiple evaluation methods to justify its superiority over multiple baselines. Finally, we wrap up with a discussion of the paper and its implications.

%We evaluate our system on multiple benchmarks in the domains of conversational dialogue and machine translation. 

\section{Recent Work}

Inspired by the continuous success of the attention mechanism \cite{bahdanau2014neural}, the Transformer network \cite{vaswani2017attention} replaces the autoregressive processing of the RNN with a global self-attention over all tokens. The architecture is divided into an encoder and decoder, which process input tokens using a set of self-attention modules.
%which distribute (pass?) contextual information between tokens. 
Each layer uses an attention module to exchange information between tokens in the sequence, while the following feed-forward layer processes this information to update the vector representation of the current token. Unlike the encoder, which allows unrestricted attention between all tokens in the sequence, the decoder uses a masked attention mechanism to allow for autoregressive generation of text (the current token cannot see future tokens).

%Cite Transformer works
%- Transformer
%- Star Transformer

% The Transformer allows for constant time information exchange between any two positions on an input sequence. While this power has allowed for impressive performance on many benchmarks, this pairwise comparison is computationally expensive, producing a running time of $O(N^2d)$ for the self-attention mechanism, where $d$ is the representation size of each token and $N$ is the number of tokens in the sequence. 
% %
% To address this problem, recently proposed modifications such as 
% %This runtime becomes increasingly difficult for longer sequences. 
% the Star Transformer \cite{guo2019star} and Sparse Transformer \cite{child2019generating} attempt to reduce computational costs by limiting the tokens available to attend to in each self-attention layer. For example, the Sparse Transformer redirects connections between tokens to reduce the complexity of self-attention to $O(n \log n)$. Other models such as  Transformer-XL \cite{dai2019transformer} attempt to reduce the computation for longer input sequences by segmenting the input into chunks and avoiding backpropagation through previous chunks. 
% %By directly providing token information from previous chunks increasingly longer sequence lengths can be processed without increasing the running time of backpropagation.

\paragraph{Hierarchical Transformers.}

%Cite hierarchical Transformer works
%- papers I recently wrote down

% While these variants of the Transformer aim to reduce the computational complexity of the self-attention (and cross-attention) mechanisms, they do not attempt to incorporate any hierarchical structure into the computation.
\newcite{liu2019hierarchical} attempt to instill a hierarchical representation by processing multiple documents and attending over them using a global attention mechanism. Similarly, \newcite{chiang2019learning} enable ranking of possible responses using a hierarchical representation of the input. However,  these models do not apply to the task of single-document response generation (the non-ranking approach used in the present paper) or other single-context generative tasks. This leaves an opportunity to provide hierarchy to these use cases.

% With the advent and use of the RNN sequence-to-sequence architecture \cite{sutskever2014sequence}, \newcite{serban2016building} speculate that improved performance can be achieved using a hierarchical, or top-down abstract view of the input sequence. As such, they construct the Hierarchical Encoder-Decoder (HRED) for the purpose of dialogue generation. Each utterance in the conversation is first converted into a fixed size representation using an utterance encoder RNN. A decoder RNN then tries to predict each next utterance in the conversation. Finally, a context encoder reads the produced utterance representations to pass a fixed representation of the context history to the decoder at each generation step, to condition on all previous utterances. Using hierarchy, they show improved performance over all previous baselines.

%Cite U-Net works and say how they do not apply to this task
%- original U-Net paper with description
%- semantic segmentation papers
%- cyclegan

\paragraph{U-Net.}

In the quest to produce a hierarchical/abstract representation, the convolutional U-Net architecture \cite{ronneberger2015u} serves as a powerful inspiration. This architecture has been applied to a number of computer vision tasks to produce a 
%wholesome 
representation of input images. For example, U-Net was used in the DeepLab v3+ architecture, which achieved state-of-the-art in semantic segmentation \cite{chen2018encoder}. Similarly, it was used to produce an alignment between unaligned image domains in CycleGAN \cite{zhu2017unpaired}, so that an image of a horse could be translated to one of a zebra and vice versa. For a given input image, the U-Net operates by downsampling the image to produce smaller more abstract representations. Each layer representation has a larger number of filters/features per pixel than the previous layer. Finally, the highest-level abstract representation undergoes progressive deconvolutions to continually upsample the image to produce an output. At each level, skip connections provide information from each down layer to the corresponding up layer of the same size. 
Therefore, U-Net stands as a strong candidate for a possible method to imbue hierarchy into the Transformer architecture.
%When imagining a method to imbue hierarchy to the Transformer architecture, the U-Net should be taken into strong consideration.

% \section{Motivation}

% In this optional section, we show diagrams of Transformer attention matrices and show that the Transformer heavily attends to utterance separators, indicating that it is striving for a place to store higher-level features. We use this to motivate the decision to add hierarchy.

% NOTE: a figure was here

\section{Model Description}

% NOTE: a figure with the architecture, moved here to appear at the same page as the model description
\begin{figure*}[ht]
\begin{center}
\includegraphics[scale=.15]{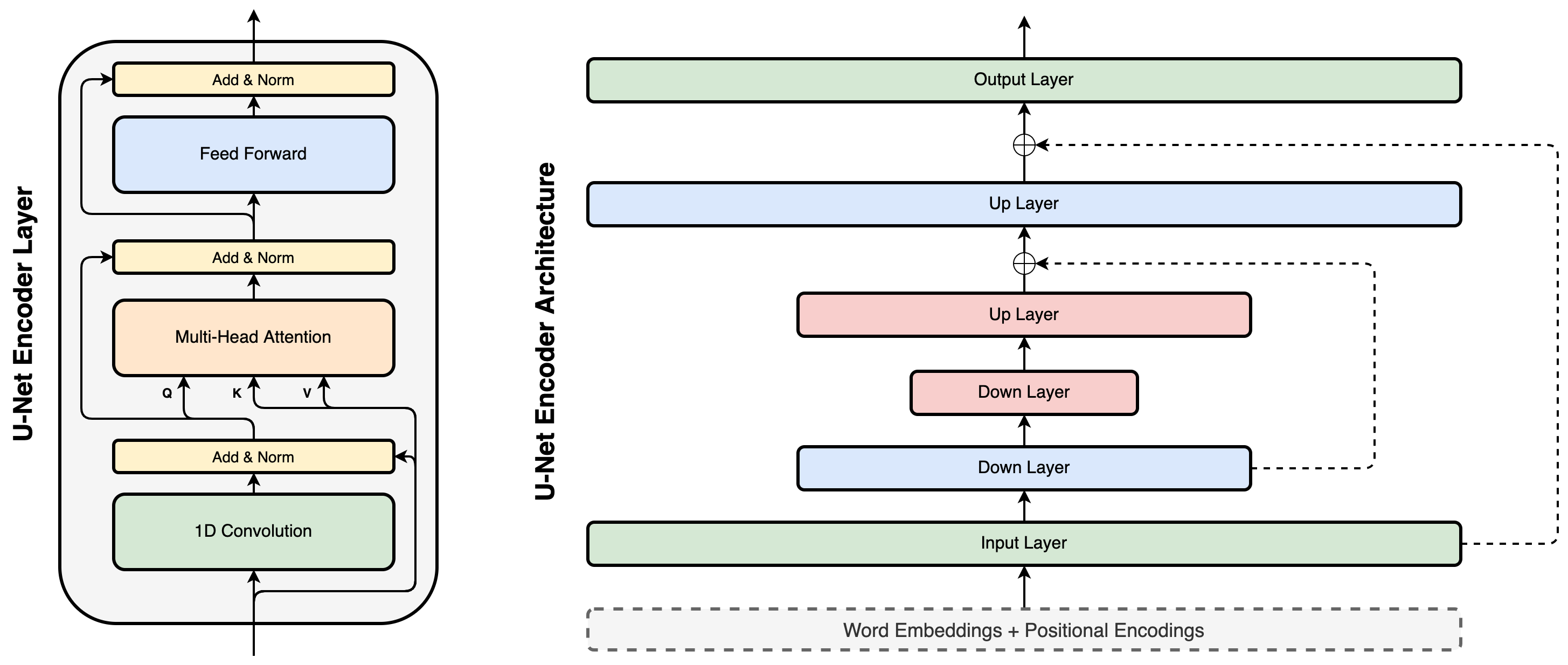}
\footnotesize
\caption{Illustration of the U-Net Transformer encoder architecture. \textbf{Left:} The U-Net encoder uses a convolution module to increase or decrease the size of each layer throughout the architecture. \textbf{Right:} The U-Net Transformer constitutes an hourglass shape which extracts high-level features in deeper layers. Residual connections combine information from abstract and low-level layers.}
\label{fig:unet}
\end{center}
\end{figure*}

%\cite{peters2018deep}

%The vanilla Transformer consists of an encoder and a decoder, each comprised by a  self-attention module and a feed-forward module in each layer \cite{vaswani2017attention}. All the layers have identical size and dimension. Thus, each layer operates on word tokens and can only perform per-token interactions via an attention mechanism. Our proposed U-Net Transformer modifies the encoder of the existing Transformer model to allow processing of the input text in a hierarchical manner. The decoder is unchanged between the two models.

The vanilla Transformer consists of an encoder and a decoder, each comprised by a self-attention module and a feed-forward module in each layer \cite{vaswani2017attention}. We propose the U-Net Transformer, which modifies the Transformer encoder to inject hierarchy into the model. The decoder and final objective are unchanged. In this section, we describe changes to the encoder.

All the layers in the vanilla Transformer have identical size and dimensions. Thus, each layer operates on word tokens and can only perform per-token interactions via the attention mechanism. 
%See Figure \ref{fig:unet} . 
% Our solution is to add the U-Net Transformer encoder layers which increase in abstraction as you move deeper in the model (see Figure \ref{fig:unet}).
We wish to produce more abstract representations that aggregate individual tokens into %abstract concept 
higher-level representations. 
To do this, we introduce convolutions into the encoder layers. This allows our novel U-Net encoder layers to change the number of token representations per layer and as well as the size of each token representation.
%To do this, we introduce novel U-Net encoder layers which can change the number of token representations per layer and change the size of each token representation through convolutions.

\paragraph{``Up'' and ``Down'' Layers.}

The architecture forms hierarchical representations in an ``hourglass'' shape  (see Figure \ref{fig:unet}). First, a series of \textbf{down} layers compress the input embeddings into increasingly abstract representations, with the number of tokens decreasing and the size of each token vector increasing. This can be thought of as modeling phrase-level components and high-level topics of the language input. Next, a series of \textbf{up} layers attempt to incorporate this abstract knowledge back into the lower-level token representations, which combine both high and low-level knowledge as input to the decoder.

To achieve this, we add a convolution module to each Transformer encoder layer before the attention module. The layer will allow us to dynamically change the size of layer outputs along both the token and embedding dimensions. This module first performs a (kernel size $k=3$) $1D$ convolution operation with a stride $1$ to aggregate information from nearby adjacent tokens.
% This convolution is selected to increase the embedding size of the output (increase the number of channels).
The convolution is followed by a linear layer to increase the embedding size of the output.
This is followed by a max pool (stride $2$) operation along the token dimension to effectively halve the number of output tokens as input to the next layer. Thus we can decrease the number of tokens and increase each token representation (in down layers). If we wish to increase the number of tokens (as in up layers), we dispose of the max pool and instead just use a deconvolution layer which maps to a larger number of output tokens with smaller embedding dimension.
% Effectively, in both up and down layers a convolution is used to control layer sizes.

\paragraph{Skip Connections.}
It is expected that an hourglass shape architecture would lose information about the input through compression in the center of the glass. To avoid losing critical information about the input, the U-Net encoder architecture features skip connections which pass information from each down layer to its corresponding up layer of the same size. We choose element-wise addition to combine both down and up representations. This technique differs from that of the computer vision U-Net in that the low and high-level representation are added instead of traditionally concatenated. This design choice was made in the same spirit as the addition of input positional and segment encodings in  the original Transformer \cite{vaswani2017attention}. It allows for combining information without increasing layer computation.

\paragraph{Attention.}

The attention module is identical to the original Transformer %implementation 
but is used to compute attention over the layer input tokens as values (pre-convolution), using the abstract (post-convolution) output tokens as queries.
This allows the attention at each layer to use high-level token information to query low-level tokens in the input. All output tokens are updated using this attention mechanism and passed to the feed-forward layer described in \newcite{vaswani2017attention}. As in the standard Transformer, a residual connection is added for each module which element-wise adds the output to the input. The convolutional residual connection (cf. Figure \ref{fig:unet}, left) is only included if the layer does not change in size. In the figure, this would be only the last layer. Layer normalization is also included after each module to normalize the activations, but does not itself change the size of the layer.

%Connecting the convolution and attention modules together, we get the following formula: 
%for layer input $x$:
Connecting the convolution and attention modules together, we compute the scaled dot product attention with modified inputs to the attention computation:
\begin{equation}
    \mbox{Attention}(Q=q, K=x, V=x)
\end{equation}
\noindent
where $x$ is the layer input and $q$ is the output of the convolution module which serves as the query:
\begin{equation}
    q = \mbox{LayerNorm}(\mbox{MaxPool}(\mbox{Conv}(x)))
\end{equation}

\paragraph{Layer Size Expansion.}
In addition to decreasing the size of each layer, the original computer vision U-Net architecture increases the size of the layer representation per-pixel (number of filters) to balance layers, so that computation is constant between layers. Specifically, the input image size halves in both width and height, causing a decrease in the number of pixels (divided by four). To compensate, the representation size (the number of filters) doubles. This is in agreement with the computational complexity of a convolutional layer $O(WHk^2C^2)$ for width $W$, height $H$, kernel size $k$ and number of filters $C$. We wish to model this same increase in layer representation in the U-Net Transformer. To accomplish this, we increase the size of each layer representation by $\sqrt{2}$ in each down layer. This is in agreement with the Transformer layer complexity of $O(Nd^2+N^2d)$ for number of input tokens $N$ and layer representation size $d$. This choice of layer size expansion limits the growth of the $Nd^2$ factor by halving $N$ and doubling $d^2$ simultaneously. See the ``Implementation'' section of the appendix for additional hyperparameter details.

% ############### CORNELL TABLE ####################

\begin{table*}[ht]
\footnotesize
\centering
\begin{tabular}{ |r|r|r| } 
 \hline
 Architecture & Lowest Validation Cross Entropy & Test Perplexity \\
 \hline
 S2SA & 3.983 & 53.43\\
 HRED & 3.945 & 51.21\\
 VHRED & 3.940 & 51.05 \\
 VHCR & 3.937 & 50.73 \\
 TRANSFORMER & 3.883 & 48.30\\
 ADJ-TRANSFORMER & 3.860 & 47.16 \\
 \hline
 UNET + LINEAR & 3.857 & 47.07 \\
 UNET & \textbf{3.853} & \textbf{46.87}\\

 \hline
\end{tabular}
\footnotesize
\caption{Comparison of U-Net Transformer against different baselines on the Cornell Movie Dialogues corpus. Model run until lowest validation loss achieved. Per-word perplexity evaluated on held-out test dataset.}
\label{table:cornell}
\end{table*}

% #################### PERSONACHAT TABLE #####################

\begin{table*}[ht]
%\small
\footnotesize
\centering
\begin{tabular}{ |r|r|r| } 
 \hline
 Architecture & Lowest Validation Cross Entropy & Test Perplexity \\

 \hline
 S2SA & 3.736 & 42.75 \\
 HRED & 3.684 & 40.48 \\
 VHCR & 3.685 & 40.45 \\
 VHRED & 3.679 & 40.11 \\
 TRANSFORMER & 3.639 & 39.06\\
 \hline
 UNET + PER-LAYER-ATT & 3.610 & 37.72 \\
 UNET & \textbf{3.581} & \textbf{36.86}\\

 \hline
\end{tabular}
\footnotesize
\caption{Comparison of U-Net Transformer against different baselines on the PersonaChat dataset. Model run until lowest validation loss achieved. Per-word perplexity evaluated on held-out test dataset.}
\label{table:personachat}
\end{table*}

\section{Baselines}

The first two baselines used in our study include a vanilla Transformer, and an RNN baseline.
%We first highlight the vanilla Transformer as the most relevant baseline of our experiments. To compare with a vanilla RNN baseline, 
For the latter, we choose the GRU sequence-to-sequence model (S2SA) with an added multi-headed attention mechanism, the same as that employed by the Transformer attention layer.

%We first highlight the vanilla Transformer as the most relevant baseline of our experiments. To compare with a vanilla RNN baseline, we choose the GRU sequence-to-sequence model (S2SA) with an added multi-headed attention mechanism, the same as that employed by the Transformer attention layer.

We also select relevant baselines from the dialogue literature. Of notable popularity are the HRED \cite{serban2016building} hierarchical architecture and the later VHRED \cite{serban2017hierarchical} model which builds on it with an added variational component. To this, we add the more recent VHCR model which claims improved performance over the HRED and VHRED models \cite{park2018hierarchical}.

\section{Evaluation}

% We choose to evaluate our proposed architecture in the domains of conversational dialogue and machine translation.

% The U-Net architecture is intended to produce explicit abstract representations of the input sequence. These abstract representations become more useful for larger input sequences which can form larger potentially larger hierarchies of information. The chit-chat generative dialogue domain is one that often requires a long conversation history in the prediction of a smaller response. We apply our proposed U-Net Transformer to this domain, to take full advantage of the modified encoder.

We evaluate our proposed model and all baselines on two dialogue datasets: the Cornell Movie Dialogues corpus \cite{Danescu-Niculescu-Mizil+Lee:11a} and the PersonaChat dataset \cite{zhang2018personalizing}. We attempt to learn from dialogue examples and generate new examples of our own.

The Cornell Movie Dialogues corpus \cite{Danescu-Niculescu-Mizil+Lee:11a} features two-character dialogues from movie scripts, and captures a large variety of human interaction in many different fictional circumstances. This dataset consists of 60,000 $\langle$history, response$\rangle$ pairs. On average, each conversation has ~3.5 previous utterances.
The PersonaChat dataset \cite{zhang2018personalizing} contains conversations that were was constructed by assigning ``personas'' to two Mechanical Turkers. A persona here is defined as a list of attributes about the individual, such as ``I love pizza.'' The Mechanical Turkers are then tasked to hold a conversation with each party adopting their assigned persona. On average, each conversation contains 14 utterances per conversation. When compared to the Cornell Movie Dialogues corpus, this dataset has longer conversation histories and thus an opportunity to form more abstract representations of the input conversation history.

% We prune the vocabularies of both datasets to no less than 20,000 tokens each; all additional tokens are replaced with the \textless unk\textgreater{}  token. To this built vocabulary we add the \textless sos\textgreater{} and \textless eos\textgreater{} tokens for denoting the beginning and end of a sequence respectively.
We use a vocabulary of the most common 20,000 tokens in each dataset.
The Transformer history input is pruned to a max of 150 tokens to put an upper bound on the memory requirements per batch.

For both datasets, we calculate cross entropy loss on the validation set as well as perplexity on a held-out test set, as a measure of how well each model is matching the language distribution. In addition, the correlation between sequence length and cross entropy loss is computed on Cornell Movie Dialogues to understand which model handles longer sequence lengths. Finally, we calculate evaluations compiled from ChatEval \cite{N19-4011} for the Cornell Movie Dialogues dataset to understand the content of output language. These include the average length of generated responses, distinct-1 and distinct-2 which capture diversity of generated output vocabulary, as well as mean and extrema scores. These embedding scores measure the  similarity between word2vec embeddings of tokens between the generated response and a correct target response from the dataset.

\section{Implementation Details}

\paragraph{Hyperparameters.}

To constrain layer computation, we set the Transformer layer size to 256 for both word embeddings and hidden dimension size. We set the fully-connected inner layer size to 1024, with attention key and value size 64 for 8 heads. As described before, the layer sizes of the U-Net Transformer change throughout the model. For a increase in size of $\sqrt{2}$ (rounded), we get layer output sizes $[256, 362, 512, 362, 256, 256]$. All key, value, and fully-connected inner sizes are proportionally scaled. The number of heads is held constant.

We similarly set the word embedding and hidden size to 256 for all baselines for both encoder and decoder (and context encoder for HRED, VHRED). The common layer size helps constrain hyperparameter tuning and is intended to allow for fair comparison of these models. The S2SA multi-headed attention mechanism has an identical structure and parameters to the Transformer attention outlined.

For both the proposed U-Net Transformer and the baseline Transformer architecture, we use 6 layers identical to the original paper \cite{vaswani2017attention}. For fair comparison with baselines, we use the Adam optimizer with learning rate $1e\minus4$ and beta parameters $(0.9, 0.999)$ without learning rate scheduling. We found learning rate scheduling did not affect model performance in the dialogue task. When generating language, we pass the conversation history to the U-Net/Transformer encoder and decode using the decoder which is identical between the two. The decoder uses greedy decoding, although other sampling methods could alternatively be used such as top-k or nucleus sampling \cite{holtzman2019curious}.

% Hyperparameters are identical across all experiments.

\paragraph{Initialization.}
We use Xavier initialization \cite{glorot2010understanding} for all parameters, with the exception of the pre-ReLU linear transformation of the feed-forward module. For this layer, we using Kaiming initialization \cite{he2015delving} as it normalizes post-ReLU activations. For the output layer of both the attention and feed-forward modules, we use Xavier initialization but multiply by a gain factor of $1/100$. In local experiments, this resulted in the best propagation of gradients from the loss %objective 
to the input %word 
embeddings. This design choice causes low output activations for these modules, and results in near-identity transformation behavior due to skip connections after each module. As such, activation values undergo fewer manipulations after each Transformer layer early in training.

\paragraph{Padding.}
The Transformer token representations are padded, so that each padding token in the input produces a zeroed representation at the output of each layer. In addition, an attention mask is computed to allow attention only between non-pad tokens in the input sequence. These changes allow for more efficient training without worrying about redundant padding tokens. For the U-Net Transformer, padding input tokens becomes non-trivial, as each layer decreases in size as a convolution over the previous layer. We compute the output padding of a U-Net Transformer down layer by performing a max pooling ($k=3$) operating over the pad vector. A given output token of the down-sampling convolution is a pad token if and only if all inputs within the kernel window are pad. Stated in the opposite way, if any input in the kernel window of a convolution is non-pad, the output will non-pad.

\paragraph{Segment Embeddings.}
In line with the original Transformer paper, we add sinusoidal position encodings to the input of the vanilla and U-Net Transformer architectures. However, in the dialogue domain, we additionally add a segmentation vector to each token to indicate which utterance in the conversation the token belongs. A different segmentation vector is learned for each utterance index. This is similar to the segment embeddings used in BERT \cite{devlin2018bert}.

% \paragraph{Machine translation}
% Training data for the WMT2014 English$\rightarrow$German dataset 
% was constructed using a \url{fairseq} preprocessing script\footnote{\url{github.com/pytorch/fairseq/tree/master/examples/translation}}, except the BPE-tokenization was done using \url{github.com/vkcom/YouTokenToMe}.

% Hyperparameters of translation model are the same as in our dialogue model. English and German vocabularies are not shared, each contains 32,000 tokens. Adam optimizer parameters and learning rate scheduling follow \cite{vaswani2017attention}. We train for 200,000 steps with batch size 128 and maximum sequence length of 50. The model is regularized with dropout $0.1$.

% Our GRU model has one layer encoder and decoder, hidden dim is 256. All other hyperparameters are the same as in transformer-based models, including multi-head attention.
% %
% Translation was performed using beam search with beam size of 3 for all translation models.

\begin{table*}[ht]
% \small
\footnotesize
\centering
\scriptsize
%\footnotesize
\begin{tabularx}{\linewidth}{ |X|X|X|X| } 
\hline
Model & History & Generated Response & Ground Truth \\
\hline

\texttt{UNET} & \texttt{hello , how are you doing ? <eos> i am doing well just watching the voice , i love it . you ?} & \texttt{i am doing great , just got done with my dog .} & \texttt{great ! i am currently working on a piece of furniture . i enjoy building it .} \\
\hline
\texttt{TRANSFORMER} & \texttt{hey , what is up ? what are you up to ? <eos> hello tell me about yourself <eos> i can beat you in chess probably} & \texttt{i am in the navy and i am in a band} & \texttt{i can beat you at cooking italian food} \\
\hline
\texttt{S2SA} & \texttt{hi ! i am excited ! i just got 50k <unk> on my youtube channel where i do makeup !} & \texttt{i am a teacher . i am a teacher .} & \texttt{cool . do you play any sports ? do you make money from your channel ?} \\
\hline
\texttt{HRED} & \texttt{hello , my father is dead} & \texttt{hello , good to talk to you . i am sorry to hear that} & \texttt{hi how are you today ?} \\
\hline
\texttt{VHRED} & \texttt{hi there ! how are you ?} & \texttt{i am good . how are you ?} & \texttt{i just acted like a superhero} \\
\hline
\texttt{VHCR} & \texttt{great ! i am currently working on a piece of furniture . i enjoy building it .} & \texttt{i love to read , i am a bit of a loner .} & \texttt{that is great i would love to be able to do that . i just swim and read mostly} \\
\hline

\end{tabularx}
\footnotesize
\caption{Samples from the U-Net Transformer and each of the baselines above for completeness. All models produce reasonable responses, but lack the diversity of the ground truth. All models trained on the PersonaChat dataset, which features individuals with different personas, or personalities.}
\label{table:examples}
\end{table*}

\section{Results}

% Perplexity results for the two dialogue datasets are shown in Tables \ref{table:cornell} and \ref{table:personachat}, respectively. Basic evaluations for the Cornell dataset can be seen in Table \ref{table:chateval}.
% %As an illustration, please see 
% Table \ref{table:examples} shows some sample responses generated by our model as well as the baselines using the PersonaChat corpus.

\begin{table*}[ht]
% \small
\centering
\footnotesize
\begin{tabular}{ |r|r|r|r|r|r| } 

 \hline
 Architecture & Avg. Length & Distinct-1 & Distinct-2 & Embedding & Extrema \\
 \hline
 HRED & 4.18 & 0.00361 & 0.00838 & 0.425 & 0.320 \\
 VHRED & 4.06 & 0.00442 & 0.0103 & 0.423 & 0.313 \\
 VHCR & 3.61 & 0.005 & 0.0122 & 0.428 & 0.315 \\
 S2SA & 8.29 & 0.230 & 0.146 & 0.577 & 0.356 \\
 TRANSFORMER & \textbf{9.49} & \textbf{0.0242} & \textbf{0.163} & 0.583 & 0.354 \\
 UNET & 9.14 & 0.0239 & 0.154 & \textbf{0.594} & \textbf{0.366} \\
 
 \hline
\end{tabular}
\footnotesize
\caption{ChatEval metrics for model-generated outputs on the Cornell Movie Dialogues corpus. Transformer produces longer sequences and is slightly better at the text diversity, while U-Net performs better on the semantic level (embedding-wise).}
\label{table:chateval}
\end{table*}

\begin{table*}[ht]
% \small
\centering
\footnotesize
\begin{tabular}{ |r|r|r|r|r|r| } 
 \hline
 Architecture & Avg. Length & Distinct-1 & Distinct-2 & Embedding & Extrema \\
 \hline
 HRED & 10.1 & 0.00597 & 0.0118 & 0.823 & 0.466 \\
 VHRED & 10.3 & 0.00675 & 0.0139 & 0.822 & 0.463 \\
 VHCR & 10.3 & 0.00675 & 0.0139 & 0.822 & 0.463 \\
 S2SA & 10.0 & 0.00675 & 0.0139 & 0.822 & 0.463 \\
 TRANSFORMER & \textbf{11.4} & \textbf{0.00932} & \textbf{0.0243} & \textbf{0.825} & 0.453 \\
 UNET & 10.8 & 0.00916 & 0.0238 & 0.818 & \textbf{0.473} \\
 
 \hline
\end{tabular}
\caption{ChatEval metrics for model-generated outputs on PersonaChat. Transformer produces longer sequences and is slightly better at the text diversity, while U-Net performs better on the embedding extrema score.}
\label{table:persona_chateval}
\end{table*}

%Here we outline the results of the proposed evaluations. 
On the Cornell Movie Dialogues, the U-Net Transformer outperforms the vanilla Transformer and all other baselines in perplexity on the held-out data as well as on mean and extrema embeddings from ChatEval mertics (cf. Table \ref{table:chateval}). Transformer outperforms U-Net by a small margin on mean output length and distinct scores.
%
%In addition (Table \ref{table:chateval}), on basic evaluations the U-Net Transformer outperformed on mean embedding and extrema embedding scores. Transformer did best on mean output length and distinct scores by a small margin.
%We believe this is due to the architectural design of the encoder and its ability to form abstract representations of the dialogue history and to utilize local information. This hypothesis is supported, as 

% talk about results on PersonaChat dataset once it is outperforming
%In addition to outperforming in the movie domain, 
The proposed U-Net Transformer also outperforms the vanilla Transformer and all other baselines for perplexity on the PersonaChat dataset. This dataset consists of everyday conversations with long dialogue histories, and serves to measure how our model performs for larger dialogue histories. See Table \ref{table:examples} for model-generated samples for this dataset. We report an ablation study of the proposed model in Table \ref{table:ablation}. We start from the U-Net Transformer (bottom), and slowly reduce it step by step to the vanilla Transformer, in order to analyze which changes resulted in performance gains. DOWN/UP represents the property that the number of tokens per layer decrease for each down layer, then increase for each up layer. CONV represents the addition of a convolutional layer. Note that without a convolutional layer, the number of tokens per layer remains constant.

%See Table \ref{table:personachat} for detailed results.

% report machine translation results

% While the dialogue domain contains long conversation histories suited for testing the hierarchical structure of the U-Net, the majority of papers instead evaluate on machine translation among other tasks.

%Table \ref{table:machinetranslation} shows the U-Net Transformer performance on the machine translation task.

% We evaluate our model against Transformer and sequence-to-sequence with attention baselines on the WMT2014 English-German dataset and report our results in Table \ref{table:machinetranslation}. We use the same model hyperparameters as in the other experiments except the vocabulary size. Although this domain is arguaby less suited to benefit from hierarchical structure of the U-Net, our model outperforms the best baseline by roughly half a BLEU point.

% report ablation study

% report about human evaluation results once they are conducted

\section{Discussion}

% discuss Cornell Movie Dialogues

\subsection{Cornell Movie Dialogues}

Performing well on the Cornell Movie Dialogues dataset is a difficult task, as dialogues come from movies where the surrounding context is not always clear. Still, we investigate how the U-Net performs relative to all other models. From our experiments, it is clear that both attention models (UNET and TRANSFORMER) outperform all previous baselines by a comparable margin. 
%This indicates that with proper training, attention layers can extract the best representations for the task. 
We also observe that the variational models outperform the non-variational models on the perplexity metric. Interestingly, the sequence-to-sequence attention model (S2SA) is outperformed by the HRED model. This is surprising, as the S2SA uses an external attention mechanism to look over input tokens and thus should have an advantage over the HRED model, which uses only a fixed-width context. The HRED does use a hierarchical pipeline. For Cornell Movie Dialogues, ChatEval metrics show that the U-Net Transformer is the best in mean embedding and extrema embedding similarity scores, but Transformer produces longer sequences and a minor increase in distinct scores (see Table \ref{table:chateval}). This indicates that Transformer outputs are slightly more varied.
%More work is needed to investigate this difference. 

%We also include UNET + LINEAR, a variant of the U-Net which transforms the output of each down layer by a linear layer before passing it through the residual connection to the corresponding up layer. Since this output is added to the output of the abstract representation from the previous up layer, this can be shown to be equivalent to concatenation and transformation of both embeddings. This operation occurs in the original U-Net, motivating its use. However, we found transformation through the residual connection did not help performance.

\subsection{PersonaChat}
% discuss personachat

The PersonaChat dataset features dialogues between Mechanical Turkers, in which both parties have access to a hidden ``persona'' which gives details about what each Turker likes and dislikes, as well as details about their life situation. As such, each dialogue is long and includes many details from a Turker's respective persona. 

%We speculate that if 
The U-Net Transformer outperforms the standard Transformer on this corpus, which may indicate that it is indeed able to extract high-level details from longer conversation histories. But the dialogue histories are simpler, so language modelling could also be a factor. 
%For the latter case, it should be noted that both 
However, note that the U-Net and vanilla Transformers share an identical decoder. 

%ChatEval metrics are shown in Table \ref{table:persona_chateval}. 
Table \ref{table:persona_chateval} ChatEval metrics.
These demonstrate similar findings to the result for Cornell Movie Dialogues in that the Transformer has longer generation lengths and diversity. The Transformer has higher mean embedding score, while U-Net Transformer receives the best extrema embedding score.

\begin{table*}[ht]
% \small
\footnotesize
\centering
\begin{tabular}{ |l|l|l| } 
 \hline
 Architecture & Test Perplexity \\
 
 \hline
TRANSFORMER & 48.30\\
% UNET - DOWN/UP - CONV & 48.92\\
TRANSFORMER + SKIP & 48.92\\
%  UNET - DOWN/UP & 47.09\\
 TRANSFORMER + SKIP + CONV & 47.09\\
 \hline
 TRANSFORMER + SKIP + CONV + DOWN/UP (UNET) & \textbf{46.87}  \\

 \hline
\end{tabular}
\footnotesize
\caption{Ablation on the U-Net architecture. SKIP represents skip-connections between constant size down and up layers. CONV represents the addition of a convolutional layer. DOWN/UP represents the property that the number of tokens per layer decrease for each down layer, then increase for each up layer. Note that without a convolutional layer, the number of tokens per layer remains constant. Conducted on the Cornell Movie Dialogues corpus.}
\label{table:ablation}
\end{table*}

\subsection{Ablation Studies}
\label{sec:ablation}

In order to determine the reasons for the performance gains, %analyzing the performance of our model on different datasets, we speculate about the reasons for our performance gains. We 
we conducted an ablation study 
%to analyze how each component of the U-Net model aids in performance, specifically 
% on the Cornell Movie Dialogues corpus (see Table \ref{table:ablation}).  

\paragraph{Skip Connections.}

Adding skip connections to the Transformer architecture (+ SKIP) seems to worsen performance. This could be that the model cannot control what information gets passed through the skip-connection, essentially averaging together earlier token representation with later ones. This could potentially be improved with a linear transformation in the skip-connection. However, these skip-connections are needed later on to avoid losing information from compression of more abstraction down/up layers.

\paragraph{Convolutions.}

Adding the convolution module to the beginning of each layer (+ CONV) helps to gather aggregate information from nearby tokens and can enforce locality on the model. It ultimately helped performance and lowered perplexity.

\paragraph{``Hourglass'' shape.}

Introducing an hourglass shape, where up/down layers increase and decrease in size ( + DOWN/UP) seems to improve performance by introducing abstract representation. This is an indication of the importance of reducing the number of input tokens while increasing layer sizes, in order to form a more abstract representation of groups of tokens. 

% We also evaluated the model when both up/down layers are removed, and convolutions are removed from each U-Net Transformer layer. Again, the removal of convolutions from this setup worsens performance, indicating that convolutions provide local connections which are more difficult to learn using the vanilla Transformer -- since the standard Transformer must utilize positional embeddings to learn increasingly sharp attention maps to grab adjacent tokens. After removing convolutions and up/down layers and convolutions, we remove skip connections to produce the vanilla Transformer, which produces the worst performance among all configurations. This demonstrates the importance of skip connections between up/down layers (or vanilla Transformer layers) for model performance.

% \begin{figure*}[ht]
% \begin{center}
% % \fbox{
% \includegraphics[scale=.4]{val_bleu.png}
% % }
% \caption{Validation BLEU curve for the WMT2014 English to German translation dataset through training. U-Net shows rapid BLEU increase early in training, indicating faster convergence.}
% \label{fig:valbleu}
% \end{center}
% \end{figure*}

\paragraph{Residual transformation.}
%In the results table for the Cornell Movie Dialogue corpus, 
We also conducted an experiment with a configuration referred to as UNET + LINEAR in the results table for the Cornell Movie Dialogue corpus (Table \ref{table:cornell}). This model is a variant of the U-Net which transforms the output of each down layer by a linear layer before passing it through the residual connection to the corresponding up layer. Since this output is added to the output of the abstract representation from the previous up layer, this can be shown to be equivalent to concatenation and transformation of both embeddings. This operation occurs in the original U-Net, motivating its use. However, we found transformation through the residual connection did not help performance.

\paragraph{Parameter count.}

In order to conduct a head-to-head comparison of U-Net Transformer and the standard Transformer, we also report the performance of a parameter-adjusted Transformer model (301-dimensional) on the Cornell Movie Dialogues corpus (shown in Table \ref{table:cornell} as ADJ-TRANSFORMER). This configuration scales the hidden size to match the parameter count of the U-Net (25M parameters, 18M for vanilla). Note that while the original Transformer matches U-Net in computational complexity, ADJ-TRANSFORMER has the same parameter count while maintaining a significantly larger number of attention comparisons, as the number of token representations does not decrease with each down layer.

Even in this setting, the U-Net still outperforms the Transformer, even though the vanilla Transformer benefits from a non-decreasing number of elements in each layer.
% Table \ref{table:machinetranslation} similarly includes a parameter-adjusted Transformer which is outperformed by our U-Net system.
In addition, the UNET - DOWN/UP from ablation Table \ref{table:ablation} measures the U-Net configuration performance where layers do not increase in size, effectively removing the parameter advantage from the U-Net. Still, this configuration outperforms both the vanilla Transformer and the parameter-adjusted variant. These findings demonstrate that parameter counts can not be used to explain improved performance of the U-Net Transformer model.
%only partially explain U-Net perplexities.

\paragraph{Encoder-Decoder Attention.}

In an additional experiment conducted on the PersonaChat dataset, we modified the cross-attention mechanism between encoder and decoder to allow each decoder layer to perform attention over each corresponding encoder layer of the same depth. We hypothesized that these connections allow the decoder direct access to abstract encoder embeddings. However, this alternative setup yielded a drop in performance relative to the U-Net architecture with pure attention over final outputs.

\subsection{Overall Findings}

Our proposed model outperforms the vanilla Transformer on the conversational dialogue tasks. The ablation study suggests that both the hierarchical structure and the local connectivity provided by convolutions aid in performance of the U-Net model. We provide evidence that combining the U-Net and Transformer architectures produces a model that better handles high-level abstraction and local dependencies. While the impact of hierarchy in the U-Net structure remains a difficult quality to evaluate, there are a number of indicators that hierarchy contributes to model performance. Table \ref{table:pearson} shows the correlation between the test loss value and the conversation history length, computed over all examples. 

\begin{table}
% \small
\footnotesize
\centering
\begin{tabular}{ |r|r| } 
 \hline
 Architecture & Length-Loss Correlation \\
 
 \hline
 TRANSFORMER & 0.0358\\
 UNET & \textbf{0.0308}\\

 \hline
\end{tabular}
\footnotesize
\caption{Pearson correlation between conversation history length and test loss. Transformer performance degrades faster than U-Net for longer sequences.}
\label{table:pearson}
\end{table}

The vanilla Transformer shows a stronger positive Pearson correlation between history length and test loss. This indicates that as history length increases, test loss increases proportionally faster for the vanilla Transformer (Pearson correlation is normalized for loss magnitude). We speculate that the U-Net is better able to utilize the conversation history, due to its top-down compressed structure of the input conversation which may scale better to longer sequences. Even though the correlation is 16\% larger for the vanilla Transformer, the overall correlation of both models is relatively weak overall, indicating there is not a strong dependence on sequence length to begin with.

%In the ablation (Table \ref{table:cornell}), we include 

%In order to conduct a head-to-head comparison of U-Net Transformer and the standard Transformer, we also report the performance of a parameter-adjusted Transformer model (301-dimensional) on the Cornell Movie Dialogues corpus (shown in Table \ref{table:cornell} as ADJ-TRANSFORMER). This configuration scales the hidden size to match the parameter count of the U-Net (25M parameters, 18M for vanilla). Note that while the original Transformer matches U-Net in computational complexity, ADJ-TRANSFORMER has 
%both 
%the same parameter count while maintaining a significantly larger number of attention comparisons, as the number of token representations does not decrease with each down layer. Even in this setting, the U-Net still outperforms the Transformer, even though the vanilla Transformer benefits from a non-decreasing number of elements in each layer. Table \ref{table:machinetranslation} similarly includes a parameter-adjusted Transformer which is outperformed by our U-Net system. In addition, the UNET - DOWN/UP from ablation Table \ref{table:ablation} measures the U-Net configuration performance where layers do not increase in size, effectively removing the parameter advantage from the U-Net. Still, this configuration outperforms both the vanilla Transformer and the parameter-adjusted variant. These findings demonstrate that parameter counts can not be used to explain improved performance of the U-Net Transformer model.
%only partially explain U-Net perplexities.

In future work, it would be interesting to apply the U-Net encoder architecture to %improve BERT performance on 
unsupervised training task in BERT. The additional U-Net layers could help the model store high-level information about the input sequence for use in predicting missing tokens, while also potentially allowing gradients to propagate easier with U-Net skip connections. We also note that the decoder was not modified in producing the U-Net encoder representations. Future work could modify this decoder to produce abstract representations of previously produced tokens in an autoregressive manner for better language modelling.

% discuss human evaluation

\section{Conclusion}

In conclusion, we introduce a novel variant of the Transformer architecture which utilizes the hierarchical and local connections of the convolutional U-Net architecture to produce an abstract representation of input sequences. We evaluate the proposed U-Net Transformer on several dialogue generation tasks and report improved performance on perplexity and certain embedding measures, while the Transformer shows increased diversity overall. Finally, we ablate different model components to assess the importance of U-Net design choices.

% If one is convinced that the optimal view of language involves concepts at a higher level than the individual tokens that make up the input sequence, we are motivated to explore architectures which form ``top-down'' representations of linguistic inputs. The movement from the static word embeddings to contextual embeddings produced by the Transformer is one such example. This work represents another step in the same research direction.

% conclude the paper with major objectives and findings, as well as future work

\bibliography{refs}
\bibliographystyle{acl_natbib}

\end{document}

\appendix

\section{Appendices}

\section{A\hspace{3mm}Additional Insights}

In addition to the conclusions of the paper, we highlight additional anecdotal findings which may be useful to the reader. It was observed over numerous experiments that the initialization of both Transformer and U-Net Transformer layers was critical to model performance.
%We found that lower initialization achieved better results.
We speculate that smaller initializations produce near-identity transformations earlier in training, and help useful input information propagate throughout all model layers.

We additionally found that perplexity values calculated throughout all experiments had low variance, with repeated runs lying within less than half a perplexity point. This finding made for reliable evaluation of the models above, and increases confidence in reported results.

\section{B\hspace{3mm}Implementation}

We use Xavier initialization \cite{glorot2010understanding} for all parameters, with the exception of the pre-ReLU  linear transformation of the feed-forward module. For this layer, we using Kaiming initialization \cite{he2015delving} as it normalizes post-ReLU activations. For the output layer of both the attention and feed-forward modules, we use Xavier initialization but multiply by a gain factor of $1/100$. In local experiments, this resulted in the best propagation of gradients from the loss objective to the input word embeddings. This design choice causes low output activations for these modules, and results in near-identity transformation behavior due to skip connections after each module. As such, activation values undergo fewer changes after each Transformer layer early in training.

The Transformer token representations are padded such that each padding token in the input produces a zeroed representation at the output of each layer. In addition, an attention mask is computed to allow attention only between non-pad tokens in the input sequence. These changes allow for more efficient training without worrying about redundant padding tokens. For the U-Net Transformer, padding input tokens becomes non-trivial, as each layer decreases in size as a convolution over the previous layer. We compute the output padding of a U-Net Transformer down layer by performing a max pooling (k=3) operating over the pad vector. A given output token of the down-sampling convolution is a pad token if and only if all inputs within the kernel window are pad. Stated in the opposite way, if any input to the kernel window of a convolution output is non-pad, the output will non-pad.

In line with the original Transformer paper, we add sinusoidal position encodings to the input of the vanilla and U-Net Transformer architectures. However, in the dialogue domain, we additionally add a segmentation vector to each token to indicate which utterance in the conversation the token belongs. A different segmentation vector is learned for each utterance index. This is similar to the segment embeddings used in BERT \cite{devlin2018bert}.

To constrain layer computation, we set the Transformer layer size to 256 for both word embeddings and hidden dimension size. We set the fully-connected inner layer size to 1024, with attention key and value size 64 for 8 heads. As described previously, the layers sizes of the U-Net Transformer change throughout the model. For a increase in size of $\sqrt{2}$ (rounded), we achieve layers sizes $[256, 362, 512, 362, 256, 256]$. All key, value, and fully-connected inner sizes are proportionally scaled. The number of heads is held constant.

We similarly set the word embedding and hidden size to 256 for all baselines for both encoder and decoder (and context encoder for HRED, VHRED). The common layer size helps constrain hyperparameter tuning and is intended to allow for fair comparison of these models. The S2SA multi-headed attention mechanism has an identical structure and parameters to the Transformer attention outlined.

\subsection{Dialogue Configuration}

For both the proposed U-Net Transformer and the baseline Transformer architecture, we use 6 layers identical to the original paper \cite{vaswani2017attention}. For fair comparison with baselines, we use the Adam optimizer with learning rate $1e\minus4$ and beta parameters $(0.9, 0.999)$ without learning rate scheduling. We found scheduling did not affect model performance in the dialogue task. We use greedy decoding, although other sampling methods could alternatively be used such as top-k or nucleus sampling \cite{holtzman2019curious}.

\subsection{Translation Configuration}

We use WMT2014 English$\rightarrow$German dataset. Training data was constructed according to this preprocessing script \footnote{\url{github.com/pytorch/fairseq/tree/master/examples/translation}} except BPE-tokenization that was done using \footnote{\url{github.com/vkcom/YouTokenToMe}}.

Hyperparameters of translation model are the same as in our dialogue model. English and German vocabularies are not shared, each contains 32,000 tokens. Adam optimizer parameters and learning rate scheduling follow \cite{vaswani2017attention}. We train for 200,000 steps with batch size 128 and maximum sequence length of 50. The model is regularized with dropout $0.1$.

Our GRU model has one layer encoder and decoder, hidden dim is 256. All other hyperparameters are the same as in transformer-based models, including multi-head attention.

Translation was performed using beam search with beam size of 3 for all translation models.

\bibliography{refs}
\bibliographystyle{acl_natbib}